\pdfoutput=1

\documentclass[11pt]{article}

\usepackage[final]{acl}

\usepackage{times}
\usepackage{latexsym}

\usepackage[T1]{fontenc}

\usepackage[utf8]{inputenc}

\usepackage{microtype}

\usepackage{inconsolata}
\usepackage{graphicx}
\usepackage{caption}
\usepackage{subcaption}
\usepackage{adjustbox}
\usepackage{float}
\title{Multi-Modal Retrieval For Large Language Model Based Speech Recognition}

\author{
    Jari Kolehmainen$^1$, 
    Aditya Gourav$^1$,
    Prashanth Gurunath Shivakumar, \\
    \textbf{Yile Gu},
    \textbf{Ankur Gandhe}, 
    \textbf{Ariya Rastrow}, 
    \textbf{Grant Strimel}, \and
    \textbf{Ivan Bulyko} \\
     \small{\href{mailto:jkolehm@amazon.com}{jkolehm@amazon.com}, \href{mailto:gouravag@amazon.com}{gouravag@amazon.com}}
}

\begin{document}
\maketitle
\begin{abstract}
Retrieval is a widely adopted approach for improving language models leveraging external information. As the field moves towards multi-modal large language models, it is important to extend the pure text based methods to incorporate other modalities in retrieval as well for applications across the wide spectrum of machine learning tasks and data types. In this work, we propose multi-modal  retrieval with two approaches: kNN-LM and cross-attention techniques. We demonstrate the effectiveness of our retrieval approaches empirically by applying them to automatic speech recognition tasks with access to external information. Under this setting, we show that speech-based multi-modal retrieval outperforms text based retrieval, and yields up to $~50\,\%$ improvement in word error rate over the multi-modal language model baseline. Furthermore, we achieve state-of-the-art recognition results on the Spoken-Squad question answering dataset.
\end{abstract}

\footnote{These two authors contributed equally to this work.}
\section{Introduction}

The wide adoption of large language models (LLMs) has driven new application areas levering this technology.
One such direction is jointly modeling multi-modal inputs and outputs with a single generative LLM model. 
In the speech domain, models such as those proposed by \citet{rubenstein2023audiopalm}, jointly model text and audio by tokenizing speech signals into discrete units.
With an expanded vocabulary encompassing tokens of multiple modalities, this modeling approach has been used in both single task~\cite{xue2023foundationtts} and multi-task settings, with \citet{maiti2023voxtlm} arguing that the multi-task training of speech-LLMs improves overall generalization of the model through synergies across tasks and modalities. 
With an expanded vocabulary encompassing tokens of multiple modalities, this modeling approach has been used in both single task~\cite{xue2023foundationtts} and multi-task settings, with \citet{maiti2023voxtlm} arguing that the multi-task training of speech-LLMs improves overall generalization of the model through synergies across tasks and modalities. 

With generalization capabilities, multi-modal LLMs such as AudioPalm~\cite{rubenstein2023audiopalm} and Seamless~\cite{barrault2023seamlessm4t}, have targeted many tasks, including Automatic Speech Recognition (ASR). These models rely on the decoder language model (LM) to generate the output text transcription when consuming tokenized speech as a prompt. Because the multi-modal LLM approach can leverage the use of a large text-only corpus for training and multi-tasking, the approach has an advantage compared to traditional ASR models such as recurrent neural network transducers (RNN-T)~\cite{makino2019recurrent} or whisper-like architectures~\cite{radford2023robust}, which rely primarily on paired audio-transcription data. However, enterprise-grade ASR systems often include further advances, such as functionality to incorporate  auxiliary information to assist decoding accuracy, which has yet to be fully addressed with these new multi-modal LLM-based approaches.

Perhaps the two most common approaches for incorporating auxiliary information for ASR have been shallow fusion with an external LM ~\cite{gourav2021personalization, zhao19d_interspeech, 9383560} and contextual biasing~\cite{sathyendra2022contextual, liu2021domain, chang2021context}. Shallow fusion with an external LM is a modular way to bias the ASR model at inference time by interpolating the probability distribution of the ASR model with that of the external LM. Shallow fusion though, can suffer from a loss of generality since the external LM does not have direct access to acoustic information. Neural biasing, meanwhile, resolves this issue by ingesting the auxiliary information directly into the acoustic model training~\cite{sathyendra2022contextual}. 
Yet, neural biasing is ultimately limited in the number of contextual documents it can ingest since attention over an ever larger number of documents renders the method less effective through dilution. We aim to address both of these limitations in this work through \emph{retrieval augmentation}.

Retrieval augmentation is a well known approach to improve existing LMs for ingesting additional information~\cite{mialon2023augmented,khandelwal2019generalization, guu2020retrieval, borgeaud2022improving,wang2023shall, zhong2022training, wu2020memformer, zhang2022gateformer, karpukhin2020dense,yasunaga2022retrieval}. The overarching idea of all retrieval augmented models is to use an external knowledge base, i.e. retrieval corpus, to improve LM performance. During inference, the retrieval corpus is queried for relevant context and information. The query usually consist of a key computed using an encoder model followed by a search step to find the closest neighbors to the key - typically in a cosine similarity or Euclidean distance sense. The retrieved neighbors are provided to the retrieval augmented LM as additional inputs, which can be used as prompts or cross-attended over.

\subsection{Contributions}

In this work, we show that using multi-modal retrieval can improve results significantly over canonical text based retrieval. Specifically, we demonstrate our method for speech recognition tasks in two settings 1) Ingesting dynamic multi-modal information; and 2) Domain adaptation of the multi-modal LLM. We propose and detail two retrieval approaches to achieve this result: a kNN-LM and a cross-attention based neural model. Experimentally, we compare each retrieval approach using two model sizes: a small model with $300$ million parameters~\cite{zhang2022opt}; and a larger  model with $7$ billion parameters. We ultimately demonstrate that while both approaches are capable of significant reduction of word error rate (WER) for domain adaptation, only the cross-attention model improved consistently speech recognition performance for the dynamic information task. We also show that multi-modal LLMs can be used effectively as key encoders for nearest neighbour search, removing the need to use an external neural model as encoder. This result leads to a deployable, application-friendly simplification which has compelling savings of compute resources.

\subsection{Related Work}
One of the first retrieval augmented LM was kNN-LM that used retrieved results to directly augment token softmax probabilities~\cite{khandelwal2019generalization}. Since the kNN-LM did not use a neural network to ingest the dynamic information, the method is easy to apply on existing models, but limited the performance compared with more involved models. 

Subsequent models such as RETRO~\cite{borgeaud2022improving} or REALM~\cite{guu2020retrieval} used a cross-attention based mechanism to incorporate the retrieved context into the causal and masked-LMs. RETRO devised a chunked cross-attention to retrieve text continuations which also allowed it to scale to a very large knowledge base, when compared with the kNN-LM. 

For speech-recognition applications, \citet{zhou2023knn} used a modified kNN-LM. In this work, the authors used a Connectionist Temporal Classification (CTC) decoder to create retrieval keys as opposed to an LM used in the standard kNN-LM~\cite{khandelwal2019generalization}. This change enabled the keys to have acoustic information, but limited the training data to consist only of transcribed speech compared with multi-modal LMs which can utilize both modalities independently. The output probabilities were computed in the same manner as in the standard kNN-LM. 
Further, retrieval methods have also been successfully applied in cold fusion~\cite{yusuf2023fly}. With a pre-trained LM as the key encoder, partial hypotheses from the decoder were used to search for text continuations, followed by contextual biasing for generating the transcription. However, the key encoder lacked phonetic context making the retrieved token accuracy low for the initial tokens and at entity start positions. In contrast, we will demonstrate that this limitation can be overcome by using a multi-modal LM instead of a pre-trained text-only LM, by incorporating the audio information into the retrieval context.

\citet{10094924} built key-value databases from semantic text and its corresponding text-to-speech (TTS) audio embeddings. The text and TTS embeddings were independently created using two different models. The retrieval database was used with approximate k-nearest neighbour search to bias the ASR model using attention. 
Meanwhile, \citet{wang2023speechtotext} experimented with a multi-modal LM and Speech2Entity retriever. The retriever, however, was not strictly multi-modal because keys were acoustic encoding of speech from a CTC model, later used to retrieve a set of textual candidate entities as values.

\section{Proposed Approach}

In the following sections, we describe the multi-modal speech-LLM and the modelling approaches to incorporate the retrieved context.

\subsection{Multi-Modal Language Models}

\begin{figure}[ht]
\begin{center}
\includegraphics[width=0.7\columnwidth]{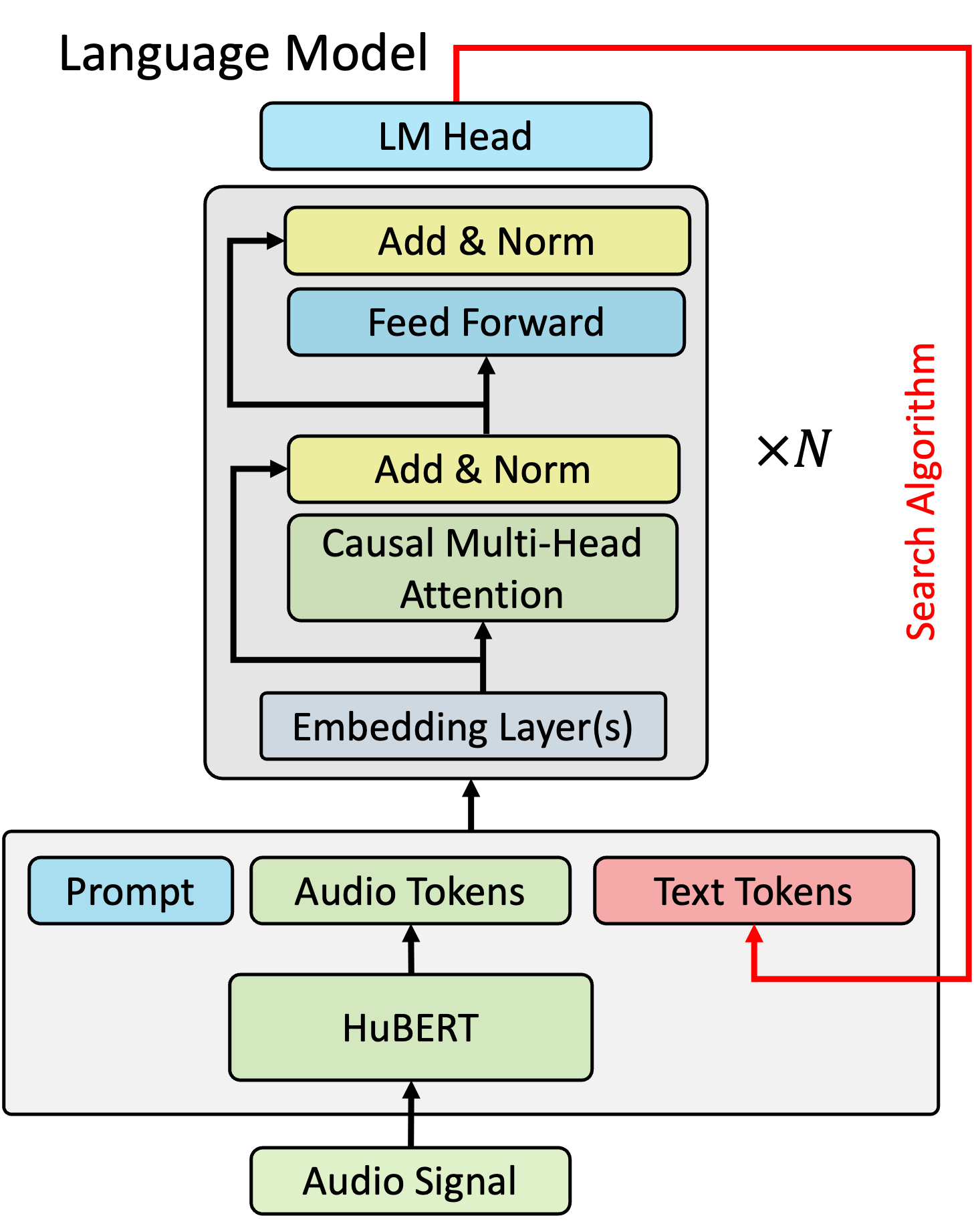}
\end{center}
\caption{Illustration of a speech multi-modal LM. Inputs to the LM consist of three parts: a prompt specifying the task, audio tokens from an audio tokenizer, and text tokens.}
\label{fig:speech-llm}
\end{figure}

Speech based multi-modal LMs model speech using quantized discrete audio tokens in addition to text tokens~\cite{rubenstein2023audiopalm}. The discrete audio tokens are extracted from pre-trained HuBERT embeddings~\cite{hsu2021hubert} followed by k-means clustering. This is illustrated in Fig. \ref{fig:speech-llm}.

\begin{figure*}[ht]
\begin{center}
\begin{subfigure}[]{1.4\columnwidth}
    \includegraphics[width=1.0\columnwidth]{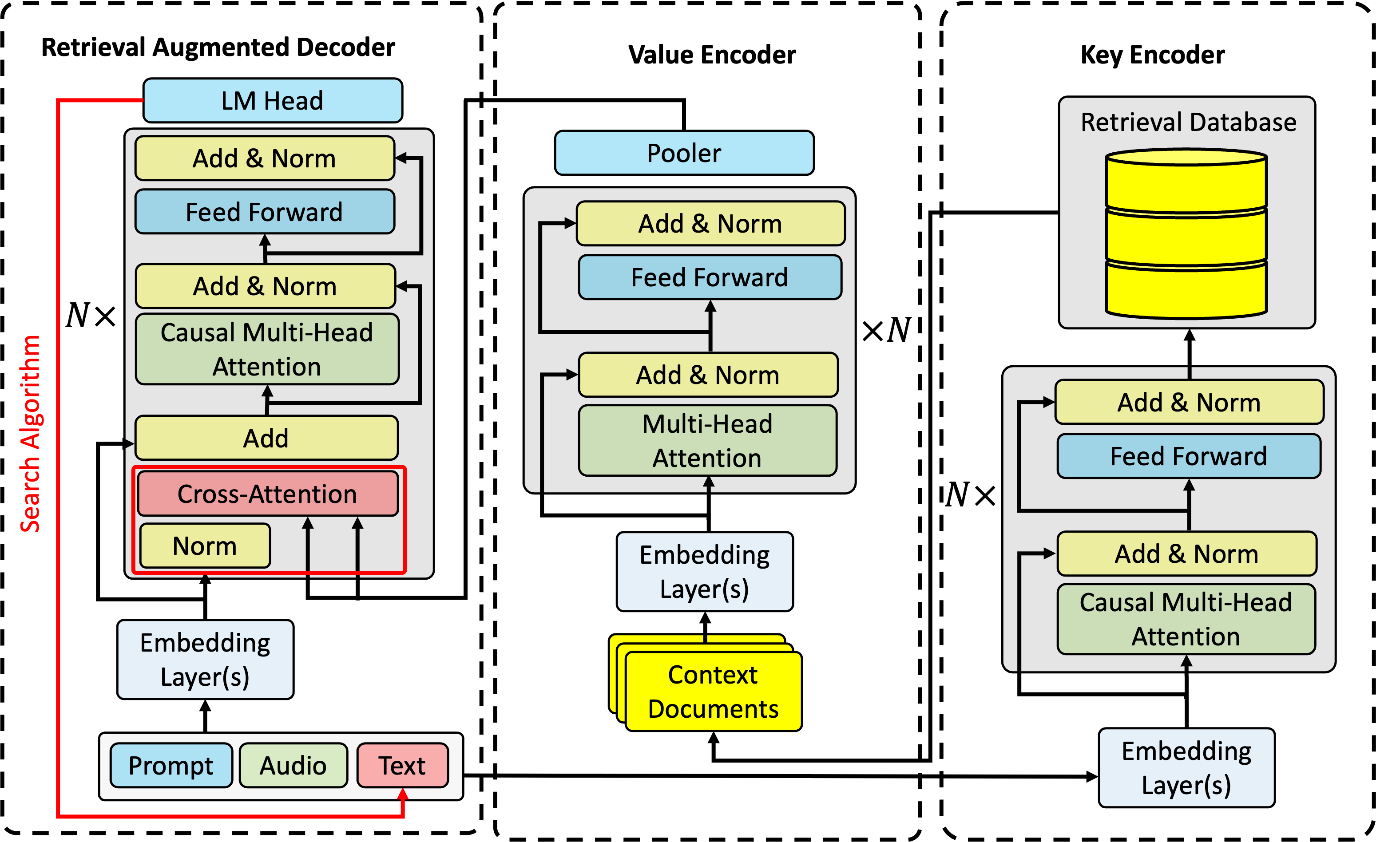}
    \caption{}
    \label{fig:illustration1}
\end{subfigure}
\begin{subfigure}[]{0.6\columnwidth}
    \includegraphics[width=1.0\columnwidth]{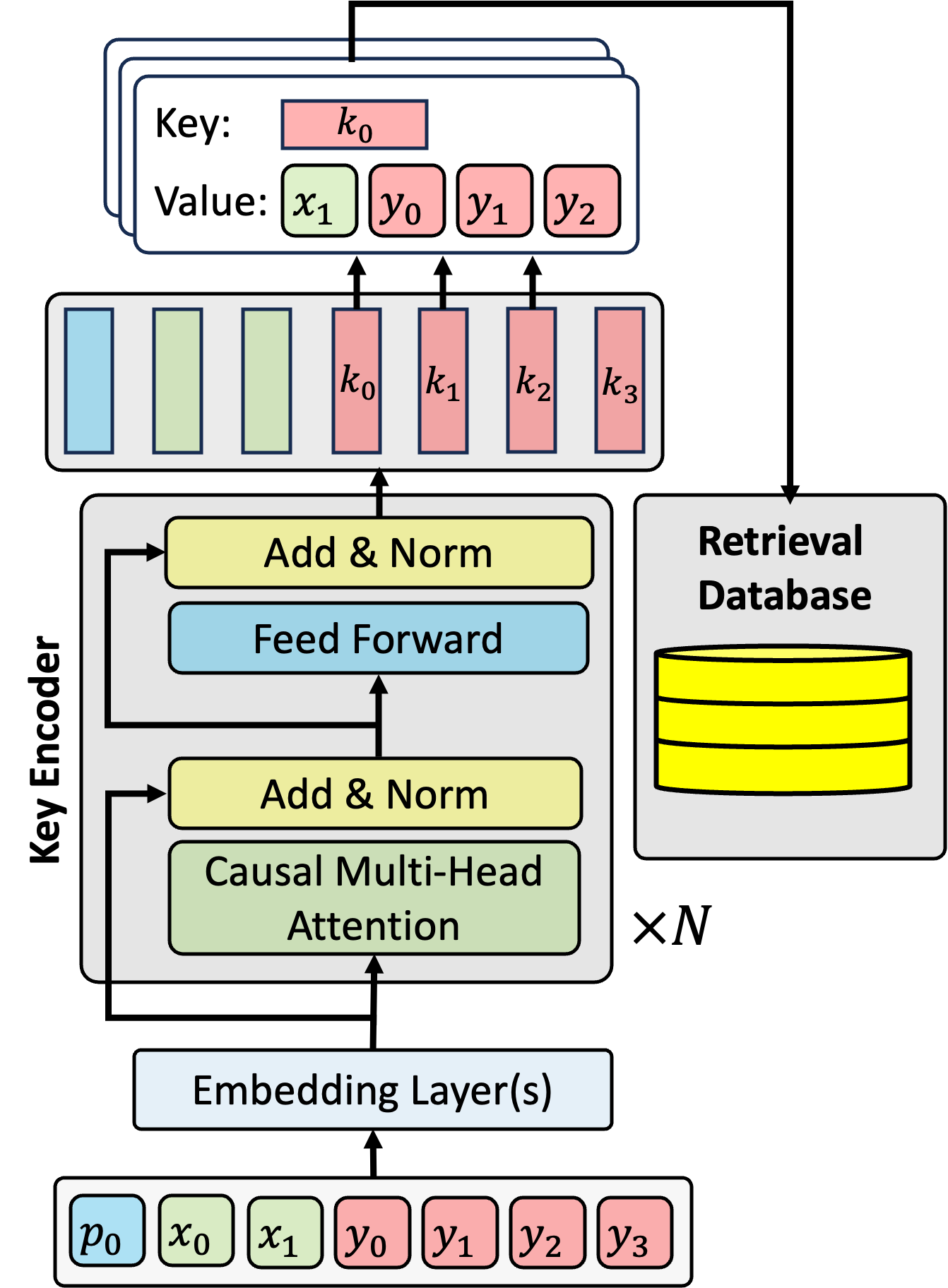}
    \caption{}
    \label{fig:database}
\end{subfigure}
\end{center}
\caption{(a): Illustration of the cross-attention based retrieval model. Input tokens are used as inputs both to the decoder model (shown on left-hand side) and key encoder (shown on the right-hand side). The encoded value are used as inputs to the decoder as the key and query for multi-head cross-attention (shown with the red-block). The depicted transformer architecture (normalization layers, etc.) is for illustration purposes and may vary slightly between different models. (b) Illustration of retrieval database creation. Text tokens are encoded and used as keys for the database. Values are surrounding tokens of the key.}
\end{figure*}

\begin{figure}[!b]
    \centering
    \includegraphics[width=\columnwidth]{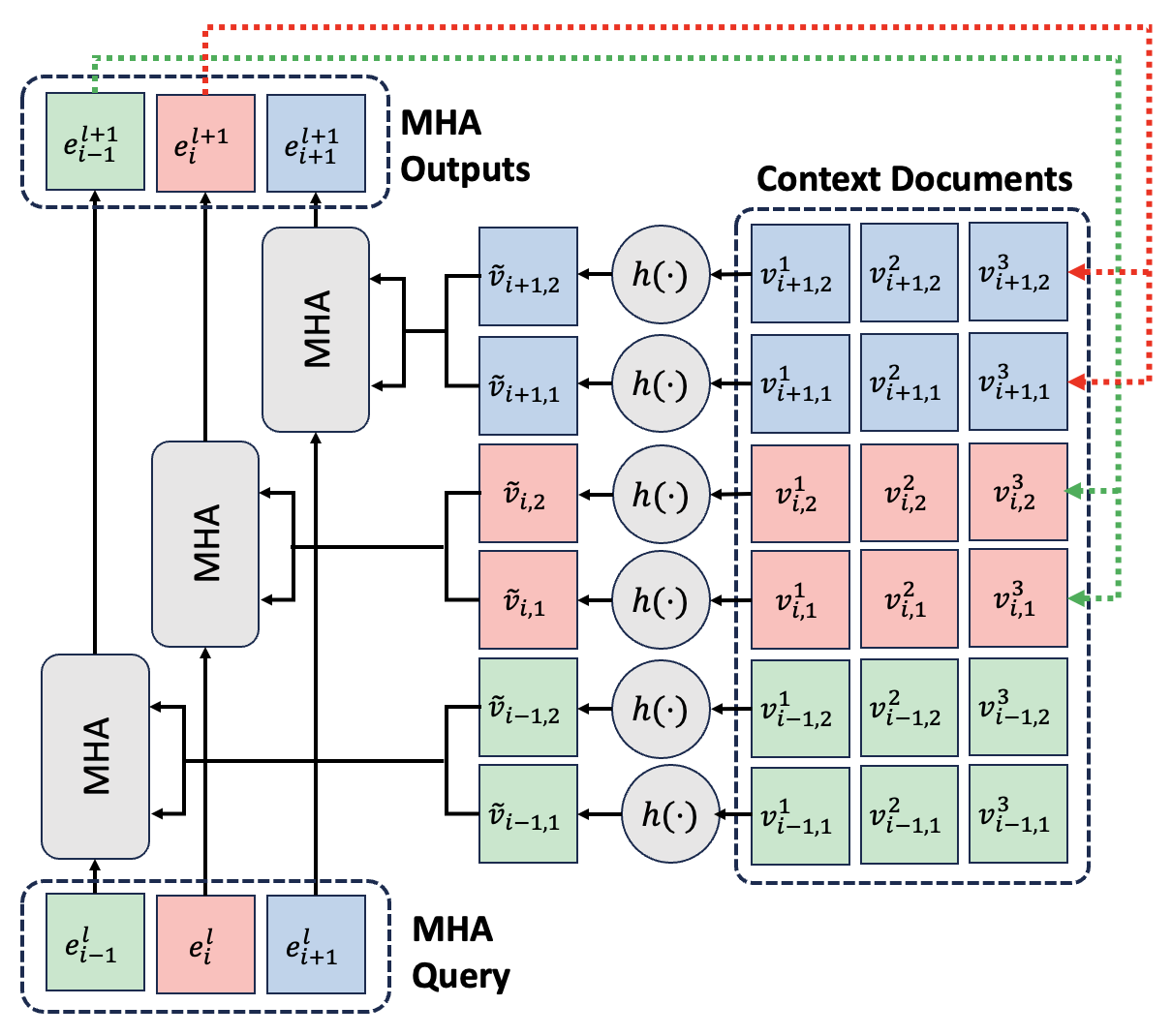}
    \caption{Illustration of the token level multi-head cross-attention. Here $e^l_{i}$ is the $i^{\mathrm{th}}$ token of the $l^{\mathrm{th}}$ layer. Color highlights the interactions between the context and the query tokens. MHA stands for standard multi-head cross-attention. The dashed arrow lines from the MHA outputs illustrate the causal dependencies.}
    \label{fig:ca_illustration}
\end{figure}

For speech recognition, the multi-modal LM is decoded by concatenating the audio tokens $x$ with a prompt $p$ and generated text tokens $y$ to form model inputs $z_{:i} = \lbrack p_0, \cdots, x_0, \cdots, y_0, \cdots,  y_{i-1} \rbrack$. Next token probability is predicted by:
\begin{equation}
P_{SLM}(y_{i}) = \mathrm{softmax}(E_o (f(z_{:i})),
\end{equation} where $P_{SLM}(y_{i})$ is the posterior distribution of the next token $y_{i}$; $f(z_{:i})$ is the multi-modal LM's last hidden state for token $y_{i-1}$; and $E_o$ is the output embedding matrix that projects the hidden state to the vocabulary dimension. The first text token - $y_0$ - is a special start-of-sentence token and the generation is continued until an end-of-sentence token is obtained or maximum sequence length is reached. In this work, the maximum sequence length is $2048$ for all models.

\subsection{Retrieval for Multi-Modal Language Models}
We consider two retrieval augmented models in this study: a kNN-LM~\cite{khandelwal2019generalization}, and a novel neural cross-attention based model. Both models aim to augment the posterior token distribution via dynamic information retrieved based on the prior tokens and follow the same retrieval search process. However, they differ in the way retrieved values are constructed and used.

\subsubsection{kNN-LM}

kNN-LM can be directly applied for speech-recognition with an exception of formatting the inputs as described in the prior section. The principal idea of kNN-LM is to directly modify the token softmax probabilities by interpolating them with a multi-modal distribution constructed from the retrieved neighbors:

\begin{equation}
    P_{kNN-LM} = \alpha P_{SLM}(y_i) + (1-\alpha) P_{kNN},
    \label{eq:knnLM}
\end{equation} where $\alpha$ is a scalar and $P_{kNN}$ is the probability distribution predicted from the retrieved neighbors given by:
\begin{equation}
   P_{kNN} \propto \sum_{j \in D(f(z_{:i}))} \mathbf{1}_{y_{i}=t_{j}} e^{-\beta \|{\bf k}_{j} - f(z_{:i})\| }.
\end{equation} Here $t_{j}$ is the retrieved next token; ${\bf k}_{j}$ is its key embedding; $D(f(z_{:i}))$ is a set containing retrieved indices; $\beta$ is a constant used to normalize the Euclidian norm; and $\mathbf{1}_{y_{i}=t_{j}}$ is the indicator function. Retrieved neighbors are obtained by finding the closest keys (in Euclidian sense) ${\bf k}_{j}$ to the models last hidden state $f(z_{:i})$. In this work, we used $\beta=10^{-3}$ for all kNN-LM experiments and $\alpha$ is found by minimizing cross-entropy in a tuning data set.

For the prompt $p$ and the audio tokens $x$ we do not perform any retrieval, and retrieve the first neighbors starting with the first generated token $y_{0}$ - a start-of-sentence token. After this token, retrieval is performed for all tokens until end-of-sentence token or maximum sequence length is reached.

\subsubsection{Cross-Attention Based Retrieval}

The cross-attention model consists of three sub-modules: 1) $f_r$ is a retrieval augmented decoder model; 2) $g$ is a key encoder model; and 3) $h$ is a value encoder model. The interplay between these models is illustrated in Fig. \ref{fig:illustration1}. The retrieval augmented decoder model is used to decode the transcription from the prior tokens (including a prompt and audio tokens). The key encoder model is used to encode all retrieval keys similar to kNN-LM while the value encoder model is used to encode the context documents.

For each token $y_{i}$ we encode the sequence $z_{:i}$ using the key encoder model $g$ to obtain a key embedding ${\bf k}_{i}$. The encoder model is a multi-modal LM that uses the same tokens and audio tokenizer as the decoder model $f_r$. The last hidden state of the multi-modal LM is used as the key embedding $k_i$ for retrieval lookup in the same way as in the kNN-LM model. 

The retrieved values $v_{i, j}$ corresponding to the key ${\bf k}_{i}$ are contiguous token sequences of fixed length extracted from a window around the key token from the corpus. The value tokens may include both speech and text tokens as opposed to individual tokens used in the kNN-LM. The retrieval database construction is illustrated in Fig. \ref{fig:database}. Each value sequence $v_{i, j}$ is encoded by the value encoder model $h$ to obtain an encoded embedding $\tilde{v}_{i,j} = h(v_{i, j})$.

In this study, we employ a small BERT~\cite{devlin2018bert} model with two transformer layers as the value encoder model (other hyper-parameters matched key encoder model $g$). Pooling the BERT embeddings is done by selecting the token embedding that follows directly after the token used for computing the corresponding key. In this study, the context document length is fixed, hence this translates to selecting a token with a predefined index. 

The encoded value vectors $\tilde{v}_{i,j}$ are stacked to form a context value matrix $\tilde{v}_i=\lbrack \tilde{v}_{i,0}, \cdots, \tilde{v}_{i,n}\rbrack$. The context value matrix is used as an input to the decoder model, $f_r$, token level cross-attention layers as the key and value matrices as illustrated in Fig. \ref{fig:ca_illustration}. More formally, for token $i$ we have:
\begin{equation}
P(y_{i}) = \mathrm{softmax}(E_o (f_r(z_{:i}, \tilde{v}_{:i}))
\end{equation}

The retrieval augmented decoder $f_r$ is constructed from a multi-modal LM by adding a token level cross-attention block with a normalization layer to selected transformer layers before the self-attention as illustrated in Fig. \ref{fig:illustration1}. In this work, we used the four topmost layers for the cross-attention blocks. During training, the original transformer weights are frozen and only the new parameters are updated. In this regard, the cross-attention layer can be considered as an adapter for new functionality.

We used pre-layer norm~\cite{xiong2020layer} for the query inputs of the cross-attention block (see Fig. \ref{fig:illustration1}) and applied an additional mask on the outputs that is constructed from the pooled context documents. If all context documents for a given token are omitted, we would zero out the output vector. This procedure in combination with the pre-layer norm and the parameter freezing guarantees that when there is no context documents provided the model predictions for all tokens matches the underlying LM.

\section{Experiments}

In the following sub-sections, we describe datasets used for experiments, model adaptation specifics, and retrieval data construction.

\subsection{Datasets}

We investigate retrieval augmentation using Spoken-Squad~\cite{lee2018spoken} and Spoken Language Understanding Evaluation (SLUE) Voxpopuli \cite{shon2022slue}. These datasets differ in utterance lengths and topics widely, allowing us to gauge the models in wide range of applications. For, Spoken-Squad, we used Amazon Polly TTS service to synthesize speech for questions and answers.

We applied a simple text normalization for all datasets: lower-casing and punctuation removal. In the case of Spoken-Squad, since the test partition was not available, we used validation partition for testing.

\subsection{Models}

We use two base multi-modal LMs in this work: a small model based on the public OPT model with 330 million parameters~\cite{zhang2022opt}; and an internal larger model using the Llama architecture~\cite{touvron2023llama} with 6.8 billion parameters. Model attributes are listed in Table \ref{tab:models}.

\begin{table}[h]
    \centering
    \caption{Summary of model hyper-parameters.}
    \label{tab:models}
    \begin{adjustbox}{width=1\columnwidth}
    \begin{tabular}{c|c|c}
        Attribute & Small & Large \\
        \hline
        Parameters & $\sim 330M$ & $\sim 6.8B$ \\
        Text tokens & $50266$ & $50001$ \\
        Speech tokens & $2000$ & $2000$ \\
        Embedding Dimension & $512$ & $4096$ \\
        Hidden Size & $1024$ & $4096$ \\
        Number of Layers & $24$ & $32$ \\
        Attention Heads &$16$& $32$ \\
        Intermediate Dimension &$4096$ & $11008$ \\
        \hline
    \end{tabular}
    \end{adjustbox}
\end{table}

Speech was encoded into continuous vectors with a pre-trained HuBERT model~\cite{hsu2021hubert} with $\sim 1$ billion parameters and further discretized using 2000 k-means centroids. Text was tokenized using the corresponding sentence piece model for all models. We used greedy search for the small model and beam search with beam width of two for the large model.

\subsection{Model Training and Fine-Tuning}

The multi-modal LMs were first pre-trained with a large text corpus. Pre-training setup for the small model is identical to \citet{zhang2022opt}. The large model was pre-trained using RedPajama~\cite{together2023redpajama} with an exception that the books subset was replaced by internal text corpus. Training hyper-parameters are obtained from \cite{touvron2023llama}. 

After text pre-training, LM vocabularies were extended with the speech tokens followed by multi-task training. The small model was trained using multi-lingual Libri-Speech~\cite{Pratap2020MLSAL} while the large model was trained using: multi-lingual Libri-Speech, Libri-Light~\cite{librilight}, People-Speech~\cite{DBLP:journals/corr/abs-2111-09344}, a large-scale multilingual speech-to-text translation corpus (CoVOST2)~\cite{wang2020covost}, Tedlium~\cite{hernandez2018ted}, and internal audio data. For all multi-tasking models, the training tasks included speech continuation, text continuation, speech recognition (ASR), and speech generation from text (TTS) with equal weights assigned. The training setup is similar to VoxtLM~\cite{maiti2023voxtlm}. Task specific prompts are listed in Table \ref{tab:format}.

\begin{table}[]
    \centering
    \begin{adjustbox}{width=1\columnwidth}
    \begin{tabular}{c|c}
        Task & Format \\
        \hline
        Text & $\langle st  \rangle$ text $\langle st  \rangle$ \\
        Speech & $\langle sa  \rangle$ audio $\langle st  \rangle$ \\
        ASR & $\langle st  \rangle$ [ASR] $\langle sa  \rangle$ audio $\langle st  \rangle$ text $\langle st  \rangle$ \\ 
        TTS & $\langle st  \rangle$ [TTS] $\langle st  \rangle$ text $\langle sa  \rangle$ audio $\langle st  \rangle$ \\
        \hline
    \end{tabular}
    \end{adjustbox}
    \caption{Token formats for pre-training tasks. Here $\langle \cdot \rangle$ are special tokens, [ASR] and [TTS] are text prompts.}
    \label{tab:format}
\end{table}

For the cross-attention models, the underlying speech multi-modal LM parameters were frozen, with only the value encoder transformer layer and cross-attention adapter blocks trained. The value encoder embedding layer was initialized with weights from the corresponding multi-modal LM input embedding layer; for the small model we used the input projection layer to up-project the embedding to the hidden size whereas for the large model input embeddings were used directly.

The cross-attention model training was divided in two stages: 1) Starting with randomized weights, the model was trained with pre-training tasks and cross-entropy loss along with context extracted from a random document. For half of the tokens, one of the random context documents contained the correct next token while the rest were incorrect. This training approach helped the model distinguish between relevant and irrelevant context. 2) Next, the model was fine-tuned using retrieved context from SLUE Voxpopuli training partition and Spoken-Squad context paragraphs. For SLUE Voxpopuli, the training partition was split in two parts: one part used for the training samples and the other part to construct the retrieval corpus (along with all the context documents from Spoken-Squad).

\subsection{Retrieval Data Construction}

Retrieval data is constructed from the corresponding train partition of the datasets with the exception of Spoken-Squad where the context paragraphs are used. Audio-transcription pairs are encoded in ASR format shown in Table \ref{tab:format} for multi-modal memory. Given an audio-text pair, the keys for each text token are the corresponding hidden states obtained after a forward pass \emph{with} the entire speech as prompt. For text-only memory, transcriptions and contexts are encoded in text format and keys are obtained in a similar way \emph{without} using speech as prompt. For both text-only and multi-modal retrieval memory, keys are encoded using the same model. Note that in both cases, the number of retrieval keys remains fixed because they correspond to the text tokens, but with the difference being whether audio was used as prompt or not. FAISS library \cite{johnson2019billion} is used for nearest neighbor search with a Voronoi based index. 

For kNN-LM, the retrieved values for a key are the corresponding next tokens and their Euclidian distance from the retrieval query. For the cross-attention model, the values included the tokens within a fixed widow, namely, seven tokens preceding the token used for creating the retrieval key and also the following eight tokens.

\subsection{Optimization}
The pre-training set-up is taken directly from the OPT and Llama papers. For the small OPT model, we used $1.0e^{-5}$ learning rate with exponential decay for training the cross-attention layers. Warm-up steps were set to 500, with a total of 10,000 steps. For the OPT model, we used 4 nodes with 8 GPUs (A10G) for training, with micro batch-size of 1 and 2 steps of gradient accumulation resulting in 64 as global batch size. For the larger Llama model, we reduced the learning rate to $5e^{-6}$ and increased the number of GPU nodes to 8 (increasing the global batch size to 128). We used bfloat-16 precision for training both models.

\section{Results and Discussion}

\subsection{Effect of Corpus Modality on Retrieval} 

We compare the effect of corpus modality using Spoken-Squad validation partition and quantify the recall of the transcription tokens in the retrieved values. 

\begin{table*}[t]
    \centering
    \caption{Token recall in Spoken-Squad validation partition. Index column shows the subsequent input token position used for recall computation relative to the input token used for the retrieval query. Document columns show the recall of subsequent retrieved tokens.} 
    \label{tab:retrieval}
    \begin{adjustbox}{width=1\textwidth}
    \begin{tabular}{c | c c | c c || c c | c c || c c }
        & \multicolumn{4}{c ||}{Small Model} & \multicolumn{4}{c ||}{Large Model} & \multicolumn{2}{c}{BERT} \\
        \hline
        & \multicolumn{2}{c|}{Speech and Text}  & \multicolumn{2}{c||}{Text} & \multicolumn{2}{c|}{Speech and Text}  & \multicolumn{2}{c||}{Text} & \multicolumn{2}{c}{Text} \\
        Index & 1st Doc. & 8 Docs. & 1st Doc. & 8 Docs. & 1st Doc. & 8 Docs. & 1st Doc. & 8 Docs. & 1st Doc. & 8 Docs. \\
        \hline
        $1$ & $69\,\%$ & $83\,\%$ & $16\,\%$ & $35\,\%$ & $85\,\%$ & $92\,\%$ & $34\,\%$ & $54\,\%$ & $10\,\%$ & $25\,\%$ \\
        $2$ & $31\%$ & $50\,\%$ & $6.8\,\%$ & $19\,\%$ & $47\,\%$ & $65\,\%$ & $16\,\%$ & $35\,\%$ & $8\%$ & $20\,\%$\\
        $3$ & $17\,\%$ & $34\,\%$ & $4.2\,\%$ & $15\,\%$ & $30\,\%$ & $49\,\%$ & $11\,\%$ & $27\,\%$ & $7\,\%$ & $19\,\%$\\
        $4$ & $12\,\%$ & $29\,\%$ & $3.9\,\%$ & $15\,\%$ & $22\,\%$ & $39\,\%$ & $9\,\%$ & $25\,\%$ & $7\,\%$ & $19\,\%$ \\
        \hline
    \end{tabular}
    \end{adjustbox}
\end{table*}

Retrieval recall statistics are shown in Table \ref{tab:retrieval}. The percentages show the fraction of the retrieved tokens matching the next token (relative to retrieval key) in the transcription. The next token is predicted with a high degree using a multi-modal corpus ($69\,\%$ and $85\,\%$) as opposed to the text-only corpus. We believe this is due to the corpus having both acoustic and semantic information. Higher number of retrieved documents increase the recall slightly for the multi-modal corpus as opposed to the text-only corpus, suggesting that lesser number of retrieved documents (and consequently compute) can be effectively used in applications, when using multi-modal memory. The subsequent tokens are predicted with significantly lower accuracy and recall than the first token, which can be attributed to the fact that LMs are trained to predict the next token but not the subsequent ones. 

This result has two implications: (1) the retrieved documents likely work well on a token level model predicting the next such as the kNN-LM; (2) models relying on chunks such as chunked cross-attention used in RETRO~\cite{borgeaud2022improving} are likely to have performance reduction when compared with token level models when the retrieved documents are obtained from a multi-modal auto-regressive LM.

We demonstrate the impact of the retrieval modality on ASR with the large multi-modal kNN-LM. For Spoken-Squad validation partition, using text-only memory for decoding results in a higher WER of $17.9\%$ when compared to the multi-modal memory case, which achieves a WER of $16.5\%$. Corresponding 1-best retrieval accuracy is $34\,\%$ and $85\,\%$ respectively as shown in \ref{tab:retrieval}. Hence, acoustic information is principally important for retrieval key construction in speech recognition applications using speech-text LLMs.

Table \ref{tab:retrieval} also shows the retrieval statistics with the bert-large-uncased model \cite{devlin2018bert}, which has similar number of parameters (336 million) as the small speech-text LM. For text-only retrieval corpus, the speech-text LM has higher recall for the next token (Index 1) for 1-best and 8-best neighbors compared to the BERT model . For subsequent tokens, BERT show a less steep decline in recall and fares better in terms of absolute recall values. When both speech and text are used for the retrieval corpus, the speech-text LM has higher 1-best and 8-best recall for all considered tokens (indices one to four).

\subsection{Speech Recognition}

We investigate our retrieval approaches in speech recognition setting. For these experiments, we only consider a multi-modal corpus for retrieval as it produced superior recall compared to the text-only corpus.

\begin{table*}[t]
    \centering
    \caption{WER on ASR datasets. Bold numbers indicate the best result obtained for the dataset. Numbers in the round brackets show WER Reduction (WERR) compared to the baseline model. Training column depicts the stage at which the datasets were used viz. speech-text LM adaptation or retrieval fine-tuning. CA stands for Cross-Attention}
    \label{tab:asr}
    \begin{adjustbox}{width=1\textwidth}
    \begin{tabular}{c|c||r|r|r||r|r|r}
        Training & Dataset & Small & kNN-LM (S) & CA (S) & Large & kNN-LM (L) & CA (L) \\
        \hline
         & Libri-Speech (clean) & $6.2$ & $3.7$ ($40$) & $\mathbf{3.4}$ ($45$) & $3.5$ & $3.7$ ($-5.7$)&  $3.6$ ($-2.9$)\\
        Adaptation & Libri-Speech (other) & $8.1$ & $10.6$ ($-30$)  & $7.9$ ($2.5$) & $\mathbf{6.5}$ & $6.8$ ($-4.6$) & $7.5$ ($-15$)\\
         & Tedlium & $16.8$ & $12.2$ ($27$)& $10.0$ ($40$) & $\mathbf{6.1}$ & $7.5 ($-23$) $ & $8.6 ($-41$)$ \\
        \hline
         & SLUE Voxpopuli & $21.4$ & $20.1$ ($6.0$) & $15.9$ ($25$) & $12.5$ & $12.4$ ($1$)& $11.3$ ($9.6$)\\
        Fine-tuning & Spoken-Squad & $27.2$ & $20.6$ ($24$)& $15.5$ ($43$)& $18.4$ & $16.5$ ($10$)& $\mathbf{8.4}$ ($54$)\\
         & Spoken-Squad (paragraph) & $27.2$ & $30.9$ ($-13$)& $16.9$ ($38$)& $18.4$ & $14.2$ ($23$) & $\mathbf{8.9}$ ($52$) \\
         \hline
    \end{tabular}
    \end{adjustbox}
\end{table*}

Table \ref{tab:asr} shows the WER evaluated on all datasets, which are grouped by their relationship to the multi-modal LMs training. Adaptation refers to the stage of adding speech-modality to the pre-trained text LM by extending its vocabulary with that of the audio tokenizer and training with ASR, speech continuation and TTS tasks. Fine-tuning refers to the training (Cross-Attention) of the speech-text LM from the Adaptation phase with retrieval. The speech adaptation datasets were not used for retrieval fine-tuning. Training partition of the datasets was used to construct the retrieval corpus. For Libri-Speech, Tedlium and SLUE-Voxpopuli, this corresponds to audio and transcription pairs and for Spoken-Squad, it is the context paragraphs' audio and text data over all Spoken-Squad titles. In the variant, Spoken-Squad (paragraph), we limit the retrieval corpus to each question/answer's corresponding context paragraph.

The small kNN-LM model show consistent improvement over the baseline with an exception of the Libri-Speech other dataset. We observe similar trend also for the large model with an exception that both Libri-Speech datasets degrade slightly. This can be explained by the strong in-domain baselines - in particular for the large model.  The WER Reduction (WERR) from kNN-LM ranges from single digits to $\sim40\,\%$ depending on the dataset and the baseline model.

For dynamic context in case of Spoken-Squad (paragraph), kNN-LM showed mixed results. For the small model we observe relative degradation of $13\,\%$, as opposed to improvement of of $23\,\%$ for the large model. This discrepancy can be attributed to underlying LMs retrieval recall (see Table \ref{tab:asr}) and the interpolation weight. In particular, re-tuning interpolation weight for the paragraph level Spoken-Squad would guarantee that the model does not degrade the baseline. 

The cross-attention (CA) model requires training and the performance is impacted by the fine-tuning data and the retrieval corpus. This aspect is magnified in the large model that has a large number of trainable parameters from the value encoder and cross-attention layers. In the case of fine-tuning data and retrieval corpus overlapping (Spoken-Squad, SLUE-Voxpopuli), the CA model performs better than the kNN-LM. For Spoken-Squad, this is most prominent where the large CA model out performs all the other models by a wide margin.

The larger improvement from CA model compared with kNN-LM can be attributed to two main factors: 1) the model has more trainable parameters (about $\sim 40M$ parameters for the small model and $~\sim 400M$ for the larger model) from the value-encoder and the additional CA blocks in the upper decoder layers; 2) kNN-LM is unable to discriminate incorrect context based on the context tokens and relies solely on the key distance while the cross-attention and value encoder allow more complex interactions between the context documents and the input tokens. With this context, one can conclude that the cross-attention approach tends to be a better candidate than kNN-LM when the dataset that we are domain adapting to is covered to some extent in the pre-training data of the model.

For dynamic context (paragraph Spoken-Squad) we see consistent improvement over the baselines from both the small and large model. Interestingly, the improvement we see in the dynamic context case is less than we observe for the whole corpus case. This finding could be attributed to two effects: 1) The CA model is fine-tuned using the whole corpus and hence might perform better using that corpus. 2) The paragraph based retrieval corpus may be missing some tokens that would be present in the whole corpus. Overall, the CA model seem a better choice for dynamic context than the kNN-LM model.

Both models demonstrate also a high tolerance for unrelated keys in the retrieval corpus. Adding unrelated data to the retrieval corpus has minimal impact or can even improve the results (see Spoken-Squad results for the whole retrieval corpus compared to paragraph level retrieval corpus) on retrieval precision or recall of the next token. This property allows domain adaptation of the kNN-LM or the cross-attention model to multiple domains by combining multiple retrieval corpus.

In Table \ref{tab:whisper} we compare our results against the large whisper v2 model~\cite{radford2023robust}  for entity heavy SLUE and Spoken-Squad datasets. As can be seen, we reach parity on SLUE Voxpopuli dataset (used in training data of whisper), and improve the WER of whisper by $\sim40\,\%$ for Spoken-Squad, achieving state-of-the-art results. 

\begin{table}[h]
    \centering
    \caption{WER Comparison of Whisper and Large Cross-Attention (CA) model on Fine-tuning datasets. WERR is shown in parentheses.}
    \label{tab:whisper}
    \begin{adjustbox}{width=1\columnwidth}
    \begin{tabular}{c|r|r}
        Dataset & Whisper & CA (L) \\
        \hline
         SLUE Voxpopuli & $\mathbf{11.2}$ & $11.3$ ($-0.8\%$)\\
         Spoken-Squad & $14.3$ & $\mathbf{8.4}$ ($41\%$)\\
         Spoken-Squad (paragraph) & $14.3$ & $\mathbf{8.9}$ ($38\%$) \\
         \hline
    \end{tabular}
    \end{adjustbox}
\end{table}

\begin{table*}[t]
\centering
\caption{Retrieved values using multi-modal and text-only memory after the model has generated \emph{"some species of"} for transcribing speech corresponding to \emph{"some species of beroe have a pair of strips of adhesive cells"}. The next token from retrieved values is highlighted with bold font. Special tokens, partial words and the speech tokens are omitted from the shown values for clarity.}
\label{tab:examples}
\begin{adjustbox}{width=1\textwidth}
\begin{tabular}{c|c}
AUDIO-TEXT MEMORY                                                        & TEXT MEMORY                                                           \\
\hline
while \textbf{be}roe preys mainly on other                                        & some species of \textbf{cy}dippids have bodies that                                \\
\hline
mouth "lips" in some species of \textbf{be}roe is a pair of narrow strips         & time – except that in two species of \textbf{the} genus ocryopsis individuals  \\
\hline
all modern ctenophores except the \textbf{be}roids have cydippid                  & either side of \textbf{the} mouth many species                                 \\
\hline
the \textbf{be}roida also known as nuda                                           & mouth "lips" in some species of \textbf{be}roe is a pair of narrow strips      \\
\hline
some species of \textbf{cy}dippids have bodies that                               & ophores as are the larvae of \textbf{some} flatworms that parasitize fish when \\
\hline
most species lack combs and the coastal \textbf{be}roids which lack tentacles and & he larvae of \textbf{some} sea anemones are parasites on                       \\
\hline
the exceptions are the \textbf{be}roids whose young are miniature be              & pods am\textbf{phi}pods and even krill                                         \\
\hline
beroe helped to mitigate the problem as \textbf{be}roe preys on other          & tenophores as are the larvae of \textbf{some} flatworms that parasitize fish \\
\hline
\end{tabular}
\end{adjustbox}

\end{table*}

Finally, in Table \ref{tab:examples} we provide qualitative example of how retrieval using multi-modal memory helps with ASR task. The correct transcription in this case is \emph{“some species of beroe have a pair of strips of adhesive cells”}. We inspect the retrieved values when the phrase \emph{“some species of”} has been generated by the model and it’s about to decode the audio frames relevant for transcribing the word \emph{“beroe”}. As can be seen in the table, recall of the word \emph{“beroe”} and its derivatives is high in retrieved values in case of the multi-modal memory, thereby helping the model to correctly transcribe the audio. We note that, without retrieval, the model transcribed the audio as \emph{"some species of burrow have a pair of strips of adhesive cells"}. We also list the retrieved values using text-only memory at the same time step for comparison.

\subsection{Inference Latency}
We explored inference latency for the small model and a retrieval database consisting of 1 million keys. In this setting, we observed mean retrieval time of around $3$-$4\,\mathrm{ms}$ for finding the $8$ closest neighbors to a key using a single CPU core. We used an inverted index based search from FAISS library with $2000$ Voronoi cells.

On the other hand, decoding ten tokens using the small OPT model from a $12\,\mathrm{s}$ audio took $260\,\mathrm{ms}$ on a single Nvidia T4 tensor core. The overall decoding latency is almost a magnitude larger than the retrieval latency which is between $30$-$40\,\mathrm{ms}$ for ten tokens. The above latency numbers depend on the hardware used, implementation details, and FAISS index used and may be different for different setups.

\section{Conclusion}

We investigate use of retrieval augmented multi-modal LMs for ingesting dynamic context and domain adaptation in speech recognition. We showed that a multi-modal LM can be effectively used for contextualizing retrieval database with audio, leading to an improvement of 10-50\% (absolute) in retrieved token accuracy and recall compared to the text-only counterpart. Furthermore, we compared masked-LM (BERT) with a similar sized multi-modal LM as key encoder for constructing a text based retrieval database. Overall, the multi-modal LM fared better for next token retrieval by 6-10\% (absolute), whereas the BERT model had better recall for subsequent tokens by 3-4\% (absolute).

We considered two different approaches for domain adaptation: a kNN-LM and a cross-attention based neural approach. Both approaches are studied using two different model sizes: a small model with $\sim300M$ parameters and a larger model with $~\sim 7B$ parameters. In all cases we used a multi-modal LM to encode the keys used for retrieval. Domain adaptation using cross-attention outperformed the kNN-LM for both small and large models by 30-100\% relative. The large multi-modal LM with cross-attention outperformed whisper model by $40\%$ relative for the entity heavy Spoken-Squad QA dataset, achieving state-of-the-art results. 

Additionally, the multi-modal LM's parameters can be shared between the key encoder and decoder. Activation from the early layers can be computed once, used for retrieval and re-used with the retrieved context in the modified top layers for retrieval augmented generation. This reduces memory footprint of the model, which can be an important consideration for embedded applications. 

Future directions of this work could be integration of more modalities (e.g video, sound etc.) into the retrieval process for improving model performance in various applications.

\section{Limitations}

We believe that the limitations of our work mainly stem from the limitations of Retrieval Augmented Generation (RAG), namely hallucinations and size of the retrieval database. With regards to hallucinations, a mismatch between the retrieval database and the task (in this case ASR) can lead to in-correct transcriptions. Homonyms and rare proper-nouns are especially prone to this. 
Additionally, size of the retrieval database is also a concern from two standpoints: 1. Having a very large database (billions of tokens) can negatively affect retrieval statistics like recall. Modeling better key encoder is one way to alleviate this. 2. Very large databases have higher memory requirements and can be costly to maintain.

In this work, we showed that for speech recognition using paired audio and text for creating retrieval database is better than using just text modality. This places a dependency on having paired audio-text data for improved performance, which might not always be readily available or accessible. In such cases, using a Text-to-Speech system can be relied upon for generating paired data but it comes with its own pitfalls viz noisy speech, inaccurate pronunciation of rare words etc. These can lead to hallucinations in the RAG process.

\bibliography{custom}

\end{document}